\documentclass[runningheads]{llncs}
\usepackage{graphicx}
\usepackage{amsmath,amssymb} 
\usepackage{color}
\usepackage{booktabs}
\usepackage{tabularx}
\usepackage{xspace}

\newcommand{\ouremph}[1]{{\textit{#1}}}

\newcolumntype{Y}{>{\centering\arraybackslash}X}
\newcolumntype{C}[1]{>{\hsize=#1cm\hsize\centering\arraybackslash}X}%
\makeatletter
\DeclareRobustCommand\onedot{\futurelet\@let@token\@onedot}
\def\@onedot{\ifx\@let@token.\else.\null\fi\xspace}
 
\def\ie{\emph{i.e}\onedot} 
 
\def\etc{\emph{etc}\onedot} 
 
\def\etal{\emph{et al}\onedot}
\makeatother

\begin{document}
\title{Can 3D Pose be Learned from \\ 2D Projections Alone?} 

\titlerunning{Can 3D Pose be Learned from \\ 2D Projections Alone?}

 \author{Dylan Drover \and 
Rohith MV\and
Ching-Hang Chen \and\\ 
Amit Agrawal \and
Ambrish Tyagi \and 
Cong Phuoc Huynh}

%
\authorrunning{D. Drover, R. MV, C. Chen, A. Agrawal, A. Tyagi, C. P. Huynh}
%

\institute{Amazon Lab126 Inc., Sunnyvale, CA, USA \\
\email{\{droverd, kurohith, chinghc, aaagrawa, \\ambrisht, conghuyn\}@amazon.com}}
\maketitle              

\begin{abstract}
\label{sect:abstract}

3D pose estimation from a single image is a challenging task in computer vision. We present a weakly supervised approach to estimate 3D pose points, given only 2D pose landmarks. Our method does not require correspondences between 2D and 3D points to build explicit 3D priors. We utilize an adversarial framework to impose a prior on the 3D structure, learned solely from their random 2D projections. Given a set of 2D pose landmarks, the generator network hypothesizes their depths to obtain a 3D skeleton. We propose a novel Random Projection layer, which randomly projects the generated 3D skeleton and sends the resulting 2D pose to the discriminator. The discriminator improves by discriminating between the generated poses and pose samples from a real distribution of 2D poses. Training does not require  correspondence between the 2D inputs to either the generator or the discriminator.
We apply our approach to the task of 3D human pose estimation. Results on Human3.6M dataset demonstrates that our approach outperforms many previous supervised and weakly supervised approaches. 

\end{abstract}

\begin{figure}[t!]
	\centering
	\includegraphics[width=1\linewidth,trim={.7cm 7.5cm 0.5cm 3.7cm},clip]{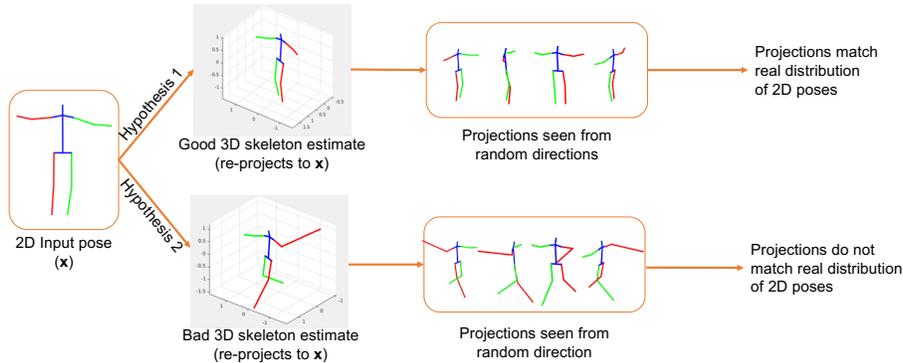}
	\caption{Key intuition behind our approach: A generator can hypothesize multiple 3D skeletons for a given input 2D pose. However, only plausible 3D skeletons will project to realistic looking 2D poses after random projections. The discriminator evaluates the ``realness'' of the projected 2D poses and provides appropriate feedback to the generator to learn to produce realistic 3D skeletons}
	\label{fig:intuition}
\end{figure}

\section{Introduction}
\label{sect:introduction}

Inferring 3D human poses from images and videos (automatic motion-capture) has garnered particular attention in the field~\cite{hogg1983model,o1980model,forsyth2006computational,moeslund2001survey} due to its numerous applications related to tracking, action understanding, human-robot-interaction and gaming, among others. Estimating 3D pose of articulated objects from 2D views is one of the long-standing ill-posed inverse problems in computer vision. We have access to, and continue to generate, large amounts of image and video data at an unprecedented rate.
This begs the question: Can we build a system that can estimate the 3D joint locations/skeleton of humans by leveraging this abundant 2D image and video data?

The problem of training end-to-end, image to 3D, pose estimation models is challenging due to variations in background, illumination, appearance, camera characteristics,~\etc. Recent approaches~\cite{MartinezICCV2017,Moreno-Noguer_2017_CVPR} have decomposed the 3D pose estimation problem into (i) estimating 2D landmark locations (corresponding to skeleton joints) and (ii) estimating 3D pose from them ({lifting 2D points to 3D}). Following such a scheme, suitable 2D pose estimators can be chosen based on the application domain~\cite{cpm,stacked-hourglass,PartAffinityCVPR2017,mask-rcnn} to estimate 2D poses, which can then be fed to a common 2D-3D lifting algorithm for recovering 3D pose.

A single 2D observation of landmarks admits infinite 3D skeletons as solution; not all these are physically plausible. The restriction of solution space to realistic poses is typically achieved by regularizing the 3D structure using priors such as symmetry, ratio of length of various skeleton elements, and kinematic constraints. These priors are often learned from ground truth 3D data, which is limited due to the complexity of capture systems. We believe that leveraging unsupervised algorithms such as generative adversarial networks for 3D pose estimation will help address the limitations of capturing such 3D data. Our work addresses the fundamental problem of lifting 2D image coordinates to 3D space without the use of any additional cues such as video~\cite{Zhou_2016_CVPR,tekin2016direct}, multi-view cameras~\cite{amin2013multi,hofmann2012multi}, or depth images~\cite{rafi2015semantic,yub2015random,shotton2013real}.

We present a weakly supervised learning algorithm to estimate 3D human skeleton from 2D pose landmarks. Unlike previous approaches we do not learn priors explicitly through 3D data or utilize explicit 2D-3D correspondence. Our system can generate 3D skeletons by only observing 2D poses.
Our paper makes the following contributions:
\begin{itemize}
	\setlength{\itemsep}{0pt}
	\setlength{\leftmargin}{0in}
	\item We present and demonstrate that a latent 3D pose distribution can be learned solely by observing 2D poses, without requiring any regression from 3D data.
	\item We propose a novel Random Projection layer and utilize it along with adversarial training to enforce a prior on 3D structure from 2D projections.
\end{itemize}

Figure~\ref{fig:intuition} outlines the key intuition behind our approach. Given an input 2D pose, there are an infinite number of 3D configurations whose projections match the position of 2D landmarks in that view. However, it is very unlikely that an implausible 3D skeleton looks realistic from another randomly selected viewpoint. On the contrary, random 2D projections of accurately estimated 3D poses are more likely to conform to the real 2D pose distribution, regardless of the viewing direction. We exploit this property to learn the prior on 3D via 2D projections. For a 3D pose estimate to be accurate (a) the projection of the 3D pose onto the original camera should be close to the detected 2D landmarks, and (b) the projection of the 3D pose onto a random camera should produce 2D landmarks that fit the distribution of real 2D landmarks.

\begin{figure}[t]
	\centering
	\includegraphics[width=1\linewidth,trim={5mm 9.5cm 7mm 5cm},clip]{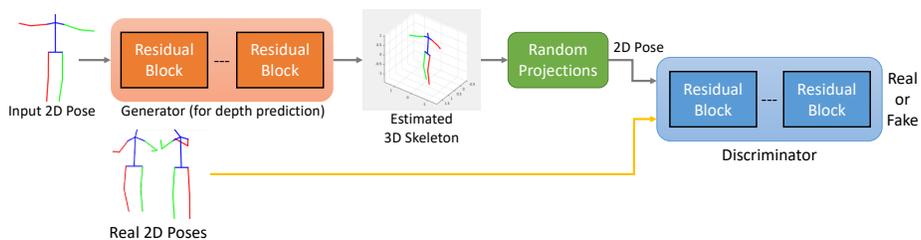}
	\caption{Adversarial training architecture for learning 3D pose}
	\label{fig:mainarch}
\end{figure}

Generative Adversarial Networks (GAN)~\cite{GAN} provide a natural framework to learn distributions without explicit supervision. Our approach learns a latent distribution (3D pose priors) indirectly via  2D poses. Given a 2D pose, the generator hypothesizes the relative depth of joint locations to obtain a 3D human skeleton. Random 2D projections of the generated 3D skeleton are fed to the discriminator, along with actual 2D pose samples (see Figure~\ref{fig:mainarch}). The 2D poses fed to the generator and discriminator do not require any correspondence during training. The discriminator learns a prior from 2D projections and enables the generator to eventually produce realistic 3D skeletons.

We demonstrate the effectiveness of our approach by evaluating 3D pose estimation on the Human3.6M dataset~\cite{h36m}. Our method shows an improvement over other weakly supervised methods which use 2D pose as input~\cite{Tung_2017_ICCV,InterpreterNetwork2016}. Interestingly we also outperform a number of supervised methods that use explicit 2D-3D correspondences.

The remainder of this paper is organized as follows. We discuss related work in Section~\ref{sect:related_work}. Section~\ref{sect:algo} provides details of the proposed method and the training methodology. Our experimental evaluation results are presented in Section~\ref{sect:experiments}. Finally, we close with concluding remarks in Section~\ref{sect:conclusions}.

\section{Related Work}
\label{sect:related_work}

\textbf{2D Pose Estimation:} Significant progress has been made recently in 2D pose estimation using deep learning techniques ~\cite{cpm,stacked-hourglass,PartAffinityCVPR2017,mask-rcnn}. Newell~\etal~\cite{stacked-hourglass} proposed a stacked hourglass architecture for predicting heatmap of each 2D joint location. Convolutional Pose Machines (CPM)~\cite{cpm} employ a sequential architecture by combining the prediction of previous stages with the input image to produce increasingly refined estimates of part locations. Cao \etal~\cite{PartAffinityCVPR2017} also estimate a part affinity field along with landmark probabilities and uses a fast, greedy search for real-time multi-person 2D pose estimation. Kaiming~\etal~\cite{mask-rcnn} accomplish this by performing fine-grained detection on top of an object detector. Our proposed method will continue to benefit from the ongoing improvement of 2D pose estimation algorithms.

\textbf{3D Pose Estimation:} Several approaches try to directly estimate 3D joint locations from images~\cite{Pavlakos_2017_CVPR,park20163d,Rogez_2017_CVPR,Towards_wild_Zhou,mehta2017vnect} in an end-to-end learning framework. However, in this paper we focus on benchmarking against methods which estimate 3D pose from 2D landmark positions, {known as} lifting from 2D to 3D~\cite{MartinezICCV2017,Tung_2017_ICCV,ChenDeva2017}. Since the input to the methods is only 2D landmark locations, it is easy to augment training of these methods using synthetic data. Like other methods in this category, our method could be combined with a variety of 2D pose estimators based on the application without retraining.
To better distinguish our work from previous approaches on lifting 2D pose landmarks to 3D, we define the following categories:

\textbf{Fully Supervised:} These include approaches such as~\cite{MartinezICCV2017,Nie_2017_ICCV,Li_2015_ICCV} that use paired 2D-3D data comprised of ground truth 2D locations of joint landmarks and corresponding 3D ground truth for learning. For example, Martinez \etal~\cite{MartinezICCV2017} learn a regression network from 2D joints to 3D joints, whereas Moreno-Noguer~\cite{Moreno-Noguer_2017_CVPR} learns a regression from a 2D distance matrix to a 3D distance matrix using 2D-3D correspondences. Exemplar based methods~\cite{ChenDeva2017,Yasin_2016_CVPR,jiang20103d} use a database/dictionary of 3D skeletons for nearest-neighbor look-up. Lin \etal~\cite{Lin_2017_CVPR} learn an end-to-end Recurrent Pose Sequence Machine whereas our approach does not use any video information. Mehta \etal~\cite{mehta2017vnect} combine a regression network which estimates 2D and 3D poses with, temporal smoothing and a parameterized, kinematic skeleton fitting method  to produce stable 3D skeletons across time. Tekin \etal~\cite{Tekin_2017_ICCV} fuse 2D and 3D image cues relying on 2D-3D correspondences. Since these methods model 3D mapping from a given dataset, they implicitly incorporate dataset-specific parameters such as camera projection matrices, distance of skeleton from camera, and scale of skeletons. This enables such models to predict metric position of joints {in 3D} on similar datasets, but requires paired 2D-3D correspondences which are difficult to obtain. 

\textbf{Weakly Supervised:} Approaches such as~\cite{Zhou_2016_CVPR,Tome_2017_CVPR,AAAI18_yxu_3dpose,Brau3DV2016} use \textit{unpaired} 3D data to learn a prior, typically as a 3D basis or articulation priors, but do not explicitly use paired 2D-3D correspondences. For example, Tome~\etal~\cite{Tome_2017_CVPR} pre-train a low-rank Gaussian model from 3D annotations as a prior for plausible 3D poses. Wu~\etal~\cite{InterpreterNetwork2016} proposed a 3D interpreter network that also estimates the weights of a 3D basis, which are learned separately for each object class using 3D data. Similarly Tung~\etal~\cite{Tung_2017_ICCV} build a 3D shape basis using PCA by aligning 3D skeletons and predicting basis coefficients. {Though this method accepts 2D landmark locations as input, this information is represented as an image within the network. On the other hand, we directly operate on the vectors of 2D landmark pixel locations, with the advantage of working in lower dimensions and avoiding convolution layers in our network.}  Zhou~\etal~\cite{Zhou_2016_CVPR} also use a 3D pose dictionary to learn pose priors. Brau~\etal~\cite{Brau3DV2016} employ an independently trained network that learns a prior distribution over 3D poses (kinematic and self-intersection priors) to impose constraints. Zhou ~\etal~ \cite{Towards_wild_Zhou} combine the 2D pose estimation task with a constraint on the bone length ratio in each skeleton group for image-to-3D pose estimation.

\textbf{Learning Using Adversarial Loss:} Recently, Generative Adversarial Networks (GAN)~\cite{GAN} have emerged as a powerful framework for learning generative models for complex data distributions. In a GAN framework, a generator is trained to synthesize samples from a latent distribution and a discriminator network is used to distinguish between synthetic and real samples. The generator's goal is to fool the discriminator by producing samples that {match the distribution of real data}. Previous approaches have used adversarial loss for human pose estimation by using a discriminator to differentiate real/fake 2D poses~\cite{Chen_2017_ICCV} and real/fake 3D poses~\cite{Tung_2017_ICCV,Black2018}. To estimate 3D, these techniques still require 3D data or use prior 3D pose model. Kanazawa~\etal~\cite{Black2018} trained an end-to-end system to estimate a skinned multi-person linear model (SMPL)~\cite{loper2015smpl} 3D mesh from RGB images by minimizing the re-projection error between 3D and 2D landmarks. However, they impose a prior on 3D skeletons using an adversarial loss with a large database of 3D human meshes. In contrast, our approach applies an adversarial loss over randomly projected 2D poses of the hypothesized 3D poses.

\section{Weakly supervised Lifting of 2D Pose to 3D Skeleton}
\label{sect:algo}

In this section we describe our weakly supervised learning approach to lift 2D human pose points to a 3D skeleton. Adversarial networks are notoriously difficult to train and we discuss design choices that lead to stable training. For consistency with generative adversarial network naming conventions, we will refer to the 3D pose estimation network as a generator. For simplicity, we work in the camera coordinate system, where the camera with unit focal length is centered at the origin $(0,0,0)$ of the world coordinate system. Let $\textbf{x}_i = (x_i,y_i), i = 1 \ldots N,$ denote $N$ 2D pose landmarks with the root joint (midpoint between hip joints) located at the origin. The 2D input pose is hence denoted by $\textbf{x} = \left[ \textbf{x}_1 \ldots \textbf{x}_N \right]$.
For numerical stability, we aim to generate 3D skeletons such that the distance from the top of the head to the root joint is approximately 1 unit.

\textbf{Generator:} The generator $\textrm{G}$ is defined as a neural network that outputs a \textit{depth offset} $o_i$ for each point $\textbf{x}_i$
\begin{equation}
\textrm{G}_{\theta_G}(\textbf{x}_i) = o_i,
\end{equation}
where $\theta_G$ are parameters of the generator learned during training. The depth of each point is defined as
\begin{equation}
z_i = \max\left(0, d + o_i \right) + 1,
\end{equation}
where $d$ denotes the distance between the camera and the 3D skeleton. Note that the choice of $d$ is arbitrary provided that $d>1$. Constraining $z_i$ to be greater than 1 ensures that the points are projected in front of the camera. In practice we use $d=10$ units.

Next, we define the back projection and the random projection layers responsible for generating the 3D skeleton and projecting it to other random views.

\textbf{Back Projection Layer:} The back projection layer takes the input 2D points $\textbf{x}_i$ and the predicted $z_i$ to compute a 3D point $\textbf{X}_i = [z_i{x}_i,z_i{y}_i,z_i]$. Note that we use exact perspective projection instead of approximations such as orthographic or paraperspective projection. 

\textbf{Random Projection Layer:} The hypothesized (generated) 3D skeleton is projected to 2D poses using randomly generated camera orientations, to be fed to the discriminator. For simplicity, we randomly rotate the 3D points (in-place) and apply perspective projection to obtain \textit{fake} 2D projections. Let $\textbf{R}$ be a random rotation matrix and $\textbf{T}=[0,0,d]$. Let $\textbf{P}_i = \left[ P^x_i, P^y_i, P^z_i \right]   = \textbf{R}(\textbf{X}_i - \textbf{T}) + \textbf{T}$ denote the 3D points after applying the random rotation. These points are re-projected to obtain fake 2D points $\textbf{p}_i = \left[ p^x_i, p^y_i \right] = \left[{P}_i^x / {P}_i^z,  {P}_i^y / {P}_i^z \right]$. The rotated points $\textbf{P}_i$ should also be in front of the camera. To ensure that, we also force ${P}_i^z$ $\geq1$. Let $\textbf{p} = \left[ \textbf{p}_1 \ldots \textbf{p}_N \right]$ denote the 2D projected pose.

Note that there is an inherent ambiguity in perspective projection; doubling the size of the 3D skeleton and the distance from the camera will result in the same 2D projection. Thus a generator that predicts absolute 3D coordinates has an additional degree of freedom between the predicted size and distance for each training sample in a batch. This could potentially result in large variance in the generator output and gradient magnitudes within a batch and cause convergence issues in training. We remove this ambiguity by predicting depth offsets with respect to a constant depth $d$ and rotating around it, resulting in stable training. In Section~\ref{sect:experiments}, we define a \textit{trivial baseline} for our approach which assumes a constant depth for all points (depth offsets equals zero, \textit{flat} human skeleton output) and show that our approach can predict meaningful depths offsets.

\textbf{Discriminator:} The discriminator $\textrm{D}$ is defined as a neural network that consumes either the fake 2D pose $\textbf{p}$ (randomly projected from generated 3D skeleton) or a real 2D pose $\textbf{r}$ (some projection, via camera or synthetic view, of a real 3D skeleton) and classifies them as either fake (target probability of 0) or real (target probability of 1), respectively.
\begin{equation}
\textrm{D}_{\theta_D} (\textbf{u})\rightarrow [0,1]
\end{equation}
where $\theta_D$ are parameters of the discriminator learned during training and $\textbf{u}$ denotes a 2D pose. Note that for any training sample $\textbf{x}$, we do {not} require $\textbf{r}$ to be same as $\textbf{x}$ or any of its multi-view correspondences. During learning we utilize a standard GAN loss~\cite{GAN} defined as
\begin{equation}
\min_G \max_D V(D,G) = \mathbb{E}(\log(D(\textbf{r}))) + \mathbb{E}(\log(1-D(\textbf{p})))
\end{equation}
Priors on 3D skeletons such as the ratio of limb lengths and joint angles are implicitly learned using only random 2D projections.

\subsection{Training}
\label{sect:training}
For training we normalize the 2D pose landmarks by centering them using the root joint and scaling the pixel coordinates so that the average head-root distance on training data is $1/d$ units in 2D. Although we can fit the entire data in GPU memory, we use a batch size of 32,768. We use the Adam optimizer~\cite{KingmaB14} with a starting learning rate of $0.0002$ for both generator and discriminator networks. We varied the batch size between 8,192 and 65,536 in experiments but it did not have any significant effect on the performance. Training time on $8$ TitanX GPUs is $0.4$ seconds per batch.

\begin{figure}[t]
	\centering
	\includegraphics[width=\linewidth,trim={.5cm 10cm 1cm 7cm},clip]{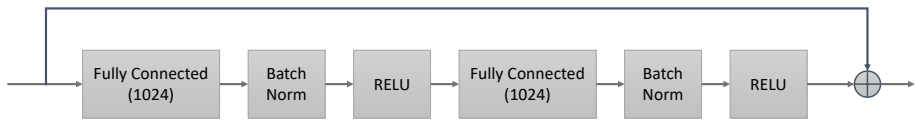}
	\caption{Residual block used in our generator and discriminator architecture}
	\label{fig:residualblock}
\end{figure}

\textbf{Generator Architecture:} The generator accepts a $28$ dimensional input representing 14 2D joint locations. Inputs are connected to a fully connected layer to expand the dimensionality to 1024 and then fed into subsequent residual blocks. Similar to ~\cite{MartinezICCV2017}, a residual block is composed of a pair of fully connected layers, each with 1024 neurons followed by batch normalization~\cite{Ioffe2015BN} and \texttt{RELU} (see Figure~\ref{fig:residualblock}). The final output is reduced through a fully connected layer to produce 14 dimensional depth offsets (one for each pose joint). A total of 4 residual blocks are employed in the generator. 

\textbf{Discriminator Architecture:} Similar to the generator, the discriminator also takes $28$ inputs representing 14 2D joint locations, either from the real 2D pose dataset or the fake 2D pose projected from the hypothesized 3D skeleton. This goes through a fully connected layer of size  1024 to feed the subsequent 3 residual blocks as defined above. Finally, the output of the discriminator is a 2-class softmax layer denoting the probability of the input being real or fake. 

\textbf{Random Rotations:} The random projection layer creates a random rotation by sampling an elevation angle $\phi$ randomly from [0,20] degrees and an azimuth angle $\theta$ from [0,360] degrees. These angles were chosen as a heuristic to roughly emulate probable viewpoints that most ``in then wild" images would have. 

\section{Experimental Results}
\label{sect:experiments}

We present quantitative and qualitative results on the widely used Human3.6M~\cite{h36m} for benchmarking. We also show qualitative visualization of reconstructed 3D skeleton from 2D pose landmarks on MPII~\cite{andriluka14cvpr} and Leeds Sports Pose~\cite{johnson2010clustered} datasets, for which the ground truth 3D data is not available.

\subsection{Dataset and Evaluation Metrics}
\label{sect:dataset_metrics}

The Human3.6M dataset is one of the largest Human Pose datasets, consisting of $3.6$ million 3D human poses. The dataset contains video and MoCap data from $5$ female and $6$ male subjects. Data is captured from $4$ different viewpoints, while subjects perform typical activities such as talking on phone, walking, eating,~\etc.

We found multiple variations of the evaluation protocols in recent literature. We report results on the two most popular protocols. Our {\bf Protocol 1} reports test results only on subject S11 to allow comparison with~\cite{ChenDeva2017,Yasin_2016_CVPR}. {\bf Protocol 2} reports results for both S9 and S11 as adopted by~\cite{MartinezICCV2017,Zhou_2016_CVPR,tekin2016direct,Tung_2017_ICCV,li20143d}. In both cases, we report the Mean Per Joint Position Error (MPJPE) in millimeters after scaling and rigid alignment to the ground truth skeleton. As discussed, our approach generates 3D skeleton up to a scale factor, since it is impossible to estimate the global scale of a human from a monocular image without additional information. Our results are based on 14-joints per skeleton.
We do not train class specific models or leverage any motion information to improve our results. The reported metrics are taken from the respective papers for comparisons.

Similar to previous works~\cite{Tung_2017_ICCV,Yasin_2016_CVPR,li20143d}, we generate synthetic 2D training data by projecting randomly rotated versions of 3D skeletons. These 2D poses are used to augment the 4 camera data already available in Human3.6M. We use additional camera positions to augment data from each 3D skeleton (we use 8 cameras compared to 144 in~\cite{Yasin_2016_CVPR}). The rotation angles for the cameras are sampled randomly in azimuth between $0$ to $360$ degrees and in elevation between $0$ to $20$ degrees. We only use data from subjects S1, S5, S6, S7, and S8 for training. 

\textbf{Trivial baseline:} We define a trivial baseline with a naive algorithm that predicts a \textit{constant} depth for each 2D pose point. This is equivalent to a generator that outputs constant depth offsets. The MPJPE of such a method is 127.3mm for \textbf{Protocol 2} using ground truth 2D points. We achieve much lower error rates in practice, reinforcing the fact that our generator is able to learn realistic 3D poses as expected.

\subsection{Quantitative Results: Protocol 1}
\label{sect:quantitative}
\begin{table*}[htb!]
\caption{Comparison of our weakly supervised approach to supervised approaches that adopt \textbf{Protocol 1}. Inputs are 2D ground truth pose points}
\label{table:protocol1_gt}
	\centering
	\begin{tabularx}{\textwidth}{ l *{8}{Y} }
			
		\toprule
		Method & Direct. & Discuss & Eat & Greet & Phone & Pose & Purchase & Sit\\
		\midrule
		Yasin~\etal~\cite{Yasin_2016_CVPR} & 60.0 & 54.7 & 71.6 & 67.5 & 63.8 & 61.9 & 55.7 & 73.9  \\
		Chen~\etal~\cite{ChenDeva2017}  &{53.3} &{46.8} &{58.6} &{61.2} &{56.0} &{58.1} &{48.9} &{55.6}\\
		Ours  &\textbf{34.3} &\textbf{36.4} &\textbf{28.4} &\textbf{33.7} &\textbf{30.0} &\textbf{43.8} &\textbf{31.7} &\textbf{32.5}\\
		\bottomrule
		\toprule
		Method  & SitDown & Smoke & Photo & Wait & Walk & WalkD & WalkP & Avg.\\
		\midrule
		Yasin~\etal~\cite{Yasin_2016_CVPR} & 110.8 & 78.9 & 96.9 & 67.9 & 47.5 & 89.3 & 53.4 & 70.5  \\
		Chen~\etal~\cite{ChenDeva2017} & {73.4} & {60.3} &{76.1} &{62.2} &{35.8} &{61.9} &{51.1} & {57.5} \\		
		Ours &\textbf{48.9}&\textbf{32.1} &\textbf{43.8} &\textbf{36.0} &\textbf{25.1} &\textbf{34.1} &\textbf{30.3} & \textbf{34.2} \\
		\bottomrule
	\end{tabularx}
\end{table*}

\begin{table*}[t]
	\centering
\caption{Comparison of our weakly supervised approach to supervised approaches that adopt \textbf{Protocol 1}. Inputs are 2D detected pose points. SH denotes stacked hourglass pose detector}
\label{table:protocol1_sh}
	\begin{tabularx}{\textwidth}{ l *{8}{Y} }
		\toprule
		Method & Direct. & Discuss & Eat & Greet & Phone & Pose & Purchase & Sit\\
		\midrule
		Yasin~\etal \cite{Yasin_2016_CVPR}& 88.4 & 72.5 & 108.5 & 110.2 & 97.1 & 81.6 & 107.2 & 119.0 \\
		Chen~\etal \cite{ChenDeva2017} &71.6 &66.6 &74.7 &79.1 &{70.1} &{67.6} &{89.3} &{90.7} \\
		Ours (SH) & \textbf{58.4}& 	\textbf{59.4}& 	\textbf{58.7}& 	\textbf{64.5}& 	\textbf{59.0} &	\textbf{60.9} &	\textbf{57.0} & \textbf{61.6}	\\

		\bottomrule
		\toprule
		Method & SitDown & Smoke & Photo & Wait & Walk & WalkD & WalkP & Avg.\\
		\midrule
		Yasin~\etal \cite{Yasin_2016_CVPR}& {170.8} & 108.2 & 142.5 & 86.9 & 92.1 & 165.7 & 102.0 & 108.3 \\
		Chen~\etal \cite{ChenDeva2017} & 195.6 & 83.5 &93.3 &71.2 &\textbf{55.7} &85.9 &62.5& 82.7 \\		
		Ours (SH) & \textbf{85.8} &	\textbf{60.4} & \textbf{64.7} & \textbf{57.4} & {63.0} & \textbf{65.5} &	\textbf{62.1} & \textbf{62.3}\\
		\bottomrule
	\end{tabularx}
\end{table*}

We first compare our approach to methods that adopt Protocol 1 in their evaluation. Table~\ref{table:protocol1_gt} compares the per class and weighted average MPJPE of our method with recent supervised learning methods~\cite{ChenDeva2017,Yasin_2016_CVPR}, using ground truth 2D points for test subject S11. Our results are superior in each category and reduces the previous error by $40\%$ (34.2mm vs. 57.5mm). Table~\ref{table:protocol1_sh} compares with the same methods using 2D points obtained from stacked hourglass(SH)~\cite{stacked-hourglass} pose detector. We similarly reduce the best reported error by $25\%$ (62.3mm vs. 82.7mm). Our method outperforms these supervised approaches in all activities, except \textit{Walking}.

\subsection{Quantitative Results: Protocol 2}
\begin{table*}[htb!]
		\caption{Comparison of our approach to other weakly supervised approaches that adopt \textbf{Protocol 2}. Inputs are 2D ground truth pose points. Results marked as $^{\ast}$ are taken from~\cite{Tung_2017_ICCV}}
	\label{table:protocol2_gt}
	\centering
		\begin{tabularx}{\textwidth}{ l *{8}{Y} }
		\toprule
		Method & Direct. & Discuss & Eat & Greet & Phone & Photo & Pose & Purchase \\
		\midrule
		3DInterpreter~\cite{InterpreterNetwork2016}$^{\ast}$ & 56.3 & 77.5 & 96.2 & 71.6 & 96.3 & 106.7 & 59.1 & 109.2\\
		Monocap~\cite{MonoCap}$^{\ast}$&  78.0 & 78.9 & 88.1 & 93.9 & 102.1 & 115.7 & 71.0 & 90.6\\
		AIGN~\cite{Tung_2017_ICCV}  & 53.7 & 71.5 & 82.3 & 58.6 & 86.9 & 98.4 & 57.6 & 104.2 \\
		Ours & {\bf 33.5}&	{\bf 39.3}&	{\bf 32.9}&	{\bf 37.0}&	{\bf 35.8}&	{\bf 42.7}&	{\bf 39.0}&	{\bf 38.2}\\

		\bottomrule
		\toprule	
		Method & Sit & SitDown & Smoke & Wait & Walk & WalkD & WalkP & Avg.\\
		\midrule
		3DInterpreter~\cite{InterpreterNetwork2016}$^{\ast}$ &111.9 & 111.9 & 124.2 & 93.3 & 58.0  & - & - & 88.6 \\
Monocap~\cite{MonoCap}$^{\ast}$ &  121.0 & 118.2 & 102.5 & 82.6 &  75.62 & - & - &  92.3  \\
AIGN~\cite{Tung_2017_ICCV} & 100.0 & 112.5 & 83.3 & 68.9 & 57.0 &  - & - &79.0 \\
Ours & {\bf 42.1}&	{\bf 52.3}&	{\bf 36.9}&	{\bf 39.4} & 	{\bf 36.8} &	{\bf 33.2}	&{\bf 34.9}&	{\bf 38.2}\\
			\bottomrule
\end{tabularx}
\end{table*}

\begin{table*}[t]
		\caption{Comparison of our approach to other weakly supervised approaches that adopt \textbf{Protocol 2}. Inputs are 2D detected pose points. SH denotes stacked hourglass. Results marked as $^{\ast}$ are taken from~\cite{Tung_2017_ICCV}}
	\label{table:protocol2_sh}
	\centering
	\begin{tabularx}{\textwidth}{ l *{8}{Y} }
		\toprule
		Method & Direct. & Discuss & Eat & Greet & Phone & Photo & Pose & Purchase \\
		\midrule
		3DInterpreter~\cite{InterpreterNetwork2016}$^{\ast}$ & 78.6 & 90.8 & 92.5 & 89.4 & 108.9 & 112.4 &  77.1 & 106.7 \\
		AIGN~\cite{Tung_2017_ICCV}  &77.6 &  91.4 &  89.9 & 88 & 107.3 & 110.1 &  75.9 &  107.5 \\
		Ours (SH) & \textbf{60.2} &	 {\bf 60.7} &	\textbf{59.2} &	{\bf 65.1} & \textbf{65.5} &	{\bf 63.8}& \textbf{59.4} & \textbf{59.4}\\
		
		\bottomrule
		\toprule	
		Method & Sit & SitDown & Smoke & Wait & Walk & WalkD & WalkP & Avg.\\
		\midrule
		3DInterpreter~\cite{InterpreterNetwork2016}$^{\ast}$&127.4 & 139.0 & 103.4 & 91.4 &  79.1 & - & - &  98.4 \\
		AIGN~\cite{Tung_2017_ICCV} & 124.2 & { 137.8} & 102.2 & 90.3 & 78.6  &  - & - &  97.2\\
		Ours (SH) & {\bf 69.1}&	\textbf{88.0}& \textbf{64.8} &	{\bf 60.8} & 	\textbf{64.9} &	\textbf{63.9}	& \textbf{65.2} &	{\bf 64.6}\\
		\bottomrule
	\end{tabularx}
\end{table*}

\begin{table}[t]
\caption{Comparison of our approach to other supervised methods on Human3.6M under \textbf{Protocol 2} using detected 2D keypoints. The results of all approaches are obtained from~\cite{MartinezICCV2017}. Our approach outperforms most supervised methods that use explicit 2D-3D correspondences}
\label{table:P2_sh_supervisedcomparison}
	\centering
	\begin{tabularx}{\textwidth}{ l *{8}{Y} }
		
		\toprule
		Method & Direct. & Discuss & Eat & Greet & Phone & Photo & Pose & Purchase \\
		\midrule
		Akhter \& Black~\cite{akhter2015pose} & 199.2 & 177.6 & 161.8 & 197.8 & 176.2 & 186.5 & 195.4 & 167.3 \\
		Ramakrishna~\etal~\cite{ramakrishna2012reconstructing} & 137.4 & 149.3 & 141.6 & 154.3 & 157.7 & 158.9 & 141.8 & 158.1 \\
		Zhou~\etal~(2016)\cite{Zhou_2016_CVPR} & 99.7 & 95.8 & 87.9 & 116.8 & 108.3 & 107.3 & 93.5 & 95.3\\
		Bogo~\etal~\cite{keep-it-simpl} & 62.0 & 60.2 & 67.8 & 76.5 & 92.1 & 77.0 & 73.0 & 75.3 \\
		Moreno-Noguer~\cite{Moreno-Noguer_2017_CVPR} & 66.1 & 61.7 & 84.5 & 73.7 & 65.2 & 67.2 & 60.9 & 67.3\\
		Martinez~\etal~\cite{MartinezICCV2017} & {44.8} & {52.0} & {44.4} & {50.5} &	{61.7} & {59.4} & {45.1} & {41.9} \\
		\midrule
		\ouremph{Ours (\textit{Weakly Supervised})} & \ouremph{60.2}&	\ouremph{60.7}&	\ouremph{59.2}&	\ouremph{65.1}&	\ouremph{65.5}&	\ouremph{63.8}&	\ouremph{59.4}&	\ouremph{59.4} \\
		\bottomrule
		
		\toprule	
		Method & Sit & SitD & Smoke & Wait & Walk & WalkD & WalkP & Avg.\\
		\midrule
		Akhter \& Black~\cite{akhter2015pose}& 160.7 & 173.7 & 177.8 & 181.9 & 176.2 & 198.6 & 192.7 & 181.1\\
		Ramakrishna~\etal~\cite{ramakrishna2012reconstructing} & 168.6 & 175.6 & 160.4 & 161.7 & 150.0 & 174.8 & 150.2 & 157.3\\
		Zhou~\etal~(2016) \cite{Zhou_2016_CVPR}&  109.1 & 137.5 & 106.0 & 102.2 & 106.5 & 110.4 & 115.2 & 106.7\\
		Bogo~\etal~\cite{keep-it-simpl}  &  100.3 & 137.3 & 83.4 & 77.3 & 86.8 & 79.7 & 87.7 & 82.3\\
		Moreno-Noguer~\cite{Moreno-Noguer_2017_CVPR} & 103.5 & 74.6 & 92.6 & 69.6 & 71.5 & 78.0 & 73.2 & 74.0\\
		Martinez~\etal~\cite{MartinezICCV2017} & {66.3} & {77.6} & {54.0} & {58.8} & {49.0} & {35.9} & {40.7} &  {52.1}\\
		\midrule
		\ouremph{Ours (\textit{Weakly supervised})} &	\ouremph{69.1}&	\ouremph{88.0}&	\ouremph{64.8}&	\ouremph{60.8}&	\ouremph{64.9}&	\ouremph{63.9}&	\ouremph{65.2}&	\ouremph{64.6}\\
		\bottomrule
	\end{tabularx}
\end{table}

\begin{table}[htb!]
\caption{Comparison of our results to the state of the art fully supervised approaches under Protocol 2 using ground truth 2D inputs. Our model has error within 1.1mm of the best supervised approach, and surpasses it with a na\"ive ensemble approach}
\label{table:fully-supervised}
\label{table:supervised}
	\small
		\centering
	\begin{tabularx}{0.97\textwidth}{ *{4}{Y} }
		\toprule
		Moreno-Noguer	& Martinez~\etal	& \multicolumn{2}{c}{Ours (Weakly supervised)} \\
		\cite{Moreno-Noguer_2017_CVPR} & \cite{MartinezICCV2017} &  { (Single Model)} & { (Ensemble)} \\
		\midrule
		62.2	& \underline{37.1} & 38.2 & {\bf 36.3} \\
		\bottomrule
\end{tabularx}
\end{table}
Next, we compare against weakly supervised approaches such as~\cite{Tung_2017_ICCV,MonoCap} that exploit 3D cues indirectly, without requiring direct 2D-3D correspondences. Table~\ref{table:protocol2_gt} compares the MPJPE for the previous weakly supervised approaches using \textbf{Protocol 2} on ground truth 2D pose inputs. Our approach reduces the error reported in~\cite{Tung_2017_ICCV} by more than $50\%$ (38.2mm vs. 79.0mm). A similar comparison is shown in Table~\ref{table:protocol2_sh} using 2D key points detected using the stacked hourglass~\cite{stacked-hourglass} pose estimator. Our approach outperforms other methods in all activity classes and reduces the previously reported error by $33\%$ (64.6mm vs. 97.2mm).

It is well known that supervised approaches broadly perform better than weakly supervised approaches in classification and regression tasks. For human 3D pose estimation, we do not expect our method to outperform the state of the art supervised approach~\cite{MartinezICCV2017}. However, our results are better than several previous published works that use 3D supervision as shown in Table~\ref{table:P2_sh_supervisedcomparison}. We have demonstrated the effectiveness of a relatively simple adversarial training framework using only 2D pose landmarks as input.

While the focus of this work is on weakly supervised learning from 2D poses alone, we are very encouraged {by the fact} that our results are competitive with the state of the art supervised approaches. In fact, our approach comes to within 1.1mm of the error reported by~\cite{MartinezICCV2017} on the ground truth 2D input as shown in Table~\ref{table:supervised}. We also experimented with a na\"ive ensemble algorithm where we combined 11 of our top performing models on the validation data and averaged the 3D skeleton for each input. This simple algorithm reduced the error to 36.3mm, surpassing the state-of-the-art results of 37.1mm (Table~\ref{table:supervised}). 

\subsection{Qualitative Results}
\begin{figure}[htb!]
	\centering
	\includegraphics[width=0.3\linewidth,trim={2.5cm .7cm 0 .7cm},clip]{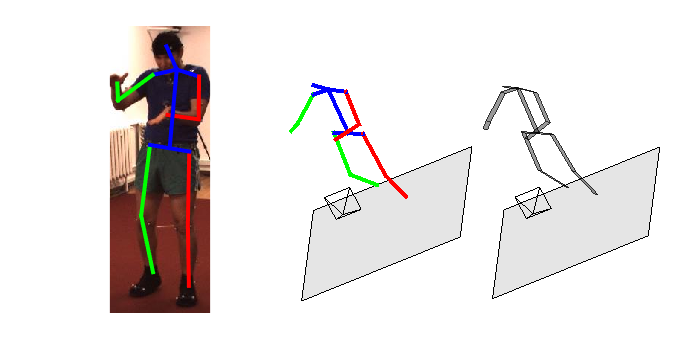}
	\includegraphics[width=0.3\linewidth,trim={2.5cm .7cm 0 .7cm},clip]{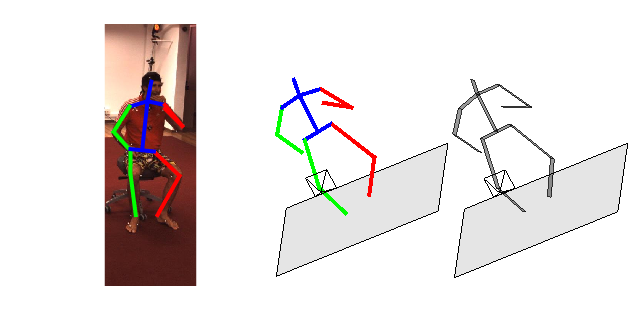}
	\includegraphics[width=0.3\linewidth,trim={2.5cm .7cm 0 .7cm},clip]{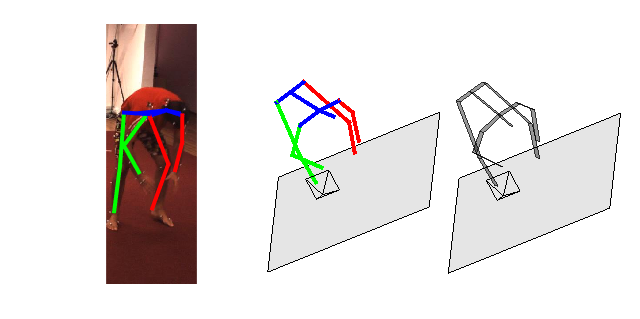}\\
	
	\includegraphics[width=0.3\linewidth,trim={2.5cm .7cm 0 .7cm},clip]{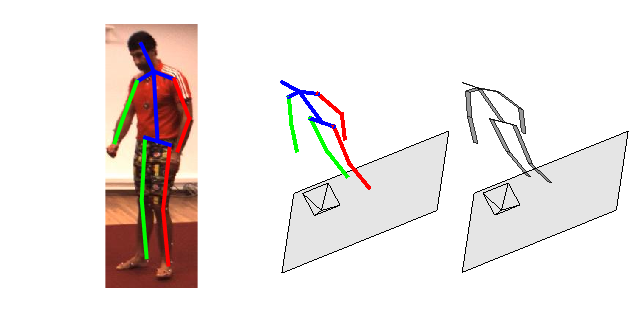}
	\includegraphics[width=0.3\linewidth,trim={2.5cm .7cm 0 .7cm},clip]{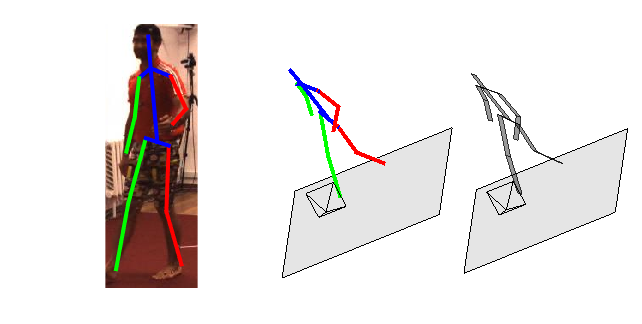}
	\includegraphics[width=0.3\linewidth,trim={2.5cm .7cm 0 .7cm},clip]{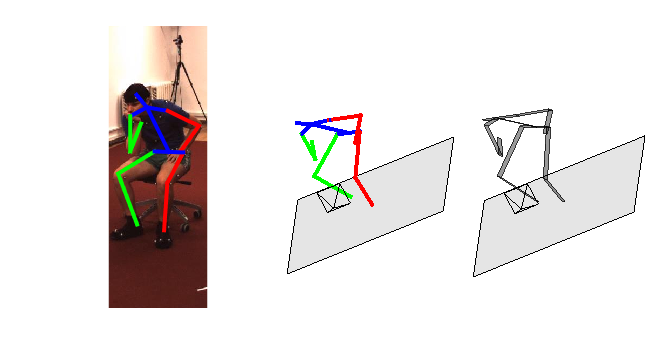}\\
	
	\includegraphics[width=0.3\linewidth,trim={2.5cm .7cm 0 .7cm},clip]{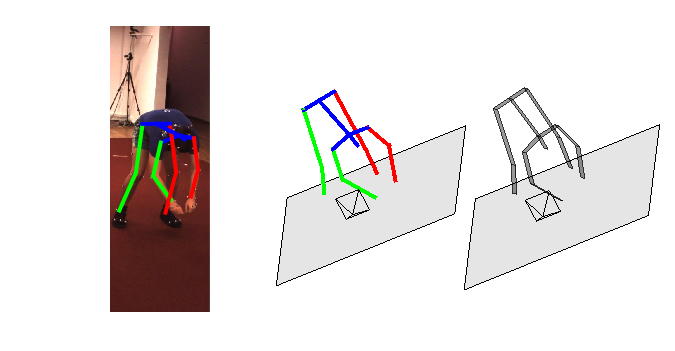}
	\includegraphics[width=0.3\linewidth,trim={2.5cm .7cm 0 .7cm},clip]{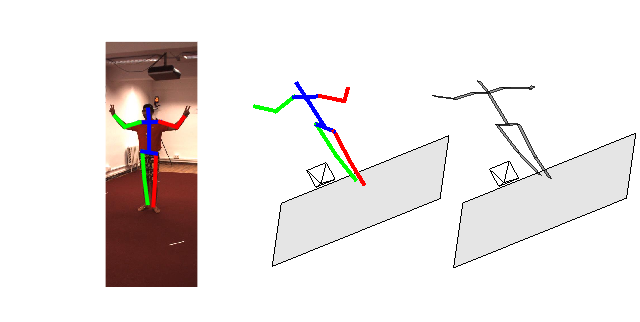}
	\includegraphics[width=0.3\linewidth,trim={2.5cm .7cm 0 .7cm},clip]{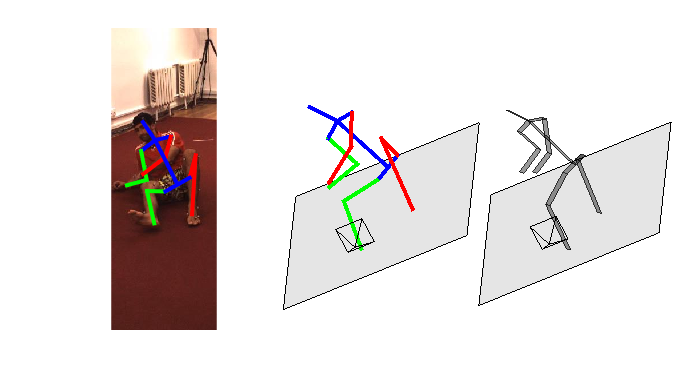}\\

	\includegraphics[width=0.3\linewidth,trim={2.5cm .7cm 0 .7cm},clip]{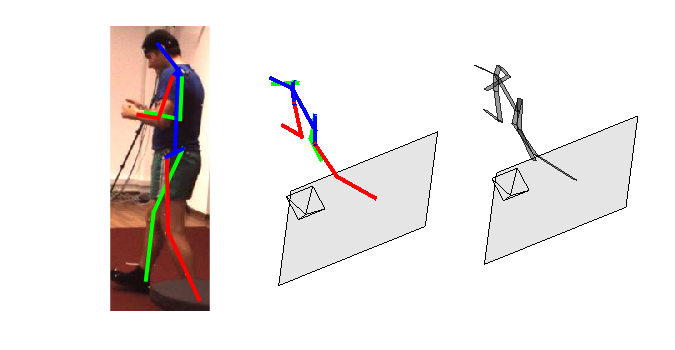}
	\includegraphics[width=0.3\linewidth,trim={2.5cm .7cm 0 .7cm},clip]{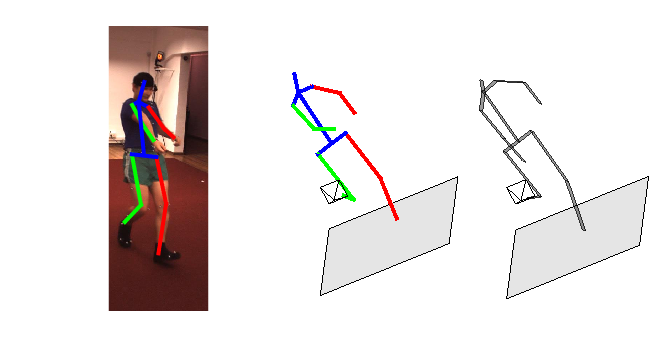}
	\includegraphics[width=0.3\linewidth,trim={2.5cm .7cm 0 .7cm},clip]{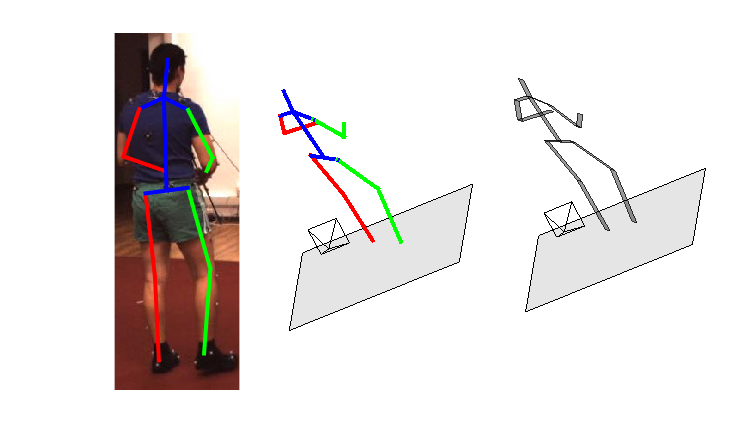}
	\\
	
	\caption{Examples of 3D pose reconstruction on Human3.6M dataset. For each image we overlay 2D pose points, followed by the predicted 3D skeleton in color. Corresponding ground truth 3D shown in gray}
	\label{fig:h36good}
\end{figure}

\begin{figure}[htb]
	\centering
\includegraphics[width=0.3\linewidth,trim={2.5cm .7cm 0 .7cm},clip]{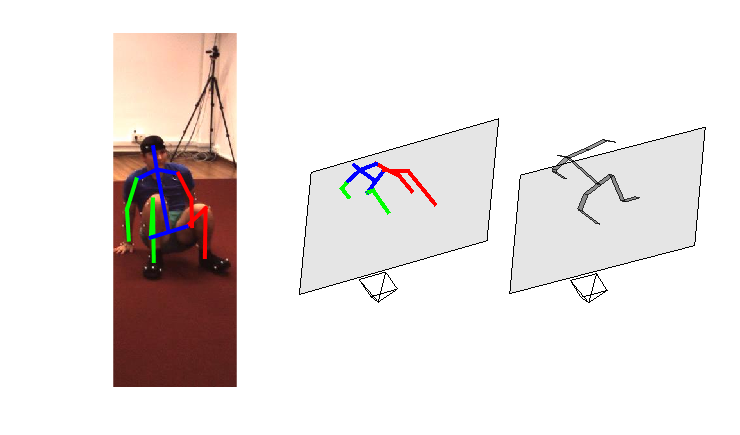}
\includegraphics[width=0.3\linewidth,trim={2.5cm .7cm 0 .7cm},clip]{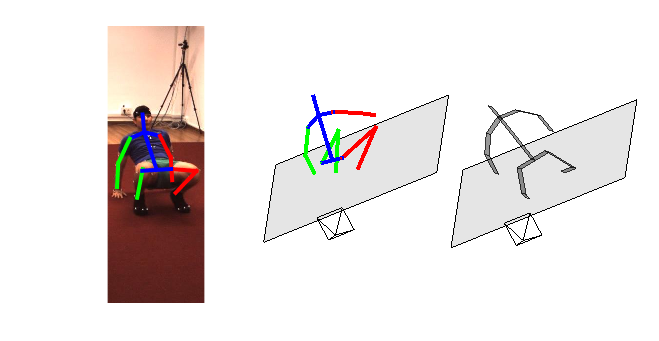}
\includegraphics[width=0.3\linewidth,trim={2.5cm .7cm 0 .7cm},clip]{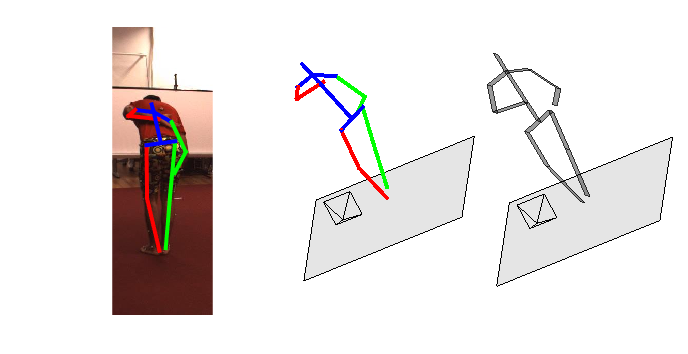} \\
\includegraphics[width=0.3\linewidth,trim={2.5cm .7cm 0 .7cm},clip]{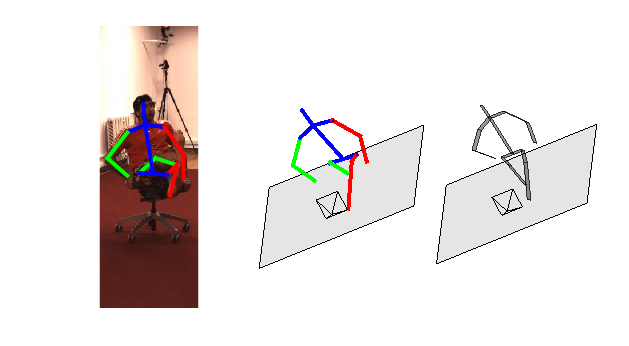}
\includegraphics[width=0.3\linewidth,trim={2.5cm .7cm 0 .7cm},clip]{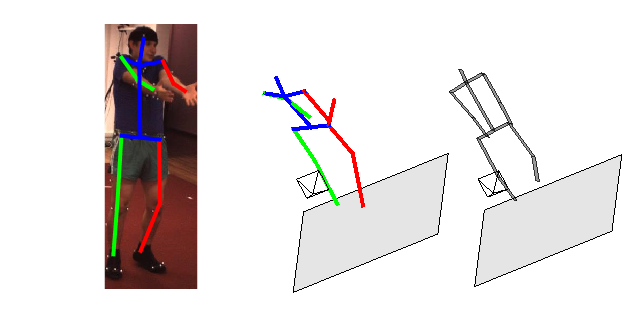}
\includegraphics[width=0.3\linewidth,trim={2.5cm .7cm 0 .7cm},clip]{images/visual_result/error_110}
	\caption{Some failure cases for our approach on Human3.6m dataset. Ground truth 3D skeleton is shown in gray}
	\label{fig:h3failure}
\end{figure}

Figure~\ref{fig:h36good} shows a few 3D pose reconstruction results on Human3.6M using our approach. The ground truth 3D skeleton is shown in gray. We see that our approach can successfully recover the 3D pose. Figure~\ref{fig:h3failure} shows some failure cases of our approach. Our typical failures are due to odd or challenging poses containing severe occlusions or plausible alternate hypothesis such as mirror flips in the direction of viewing. Since the training was performed on images containing all 14 joints, we are currently unable to lift 2D poses with fewer joints to 3D skeletons.

\begin{figure}[htb!]
	\centering
	\includegraphics[width=0.24\linewidth,trim={2.5cm .7cm 0 .7cm},clip]{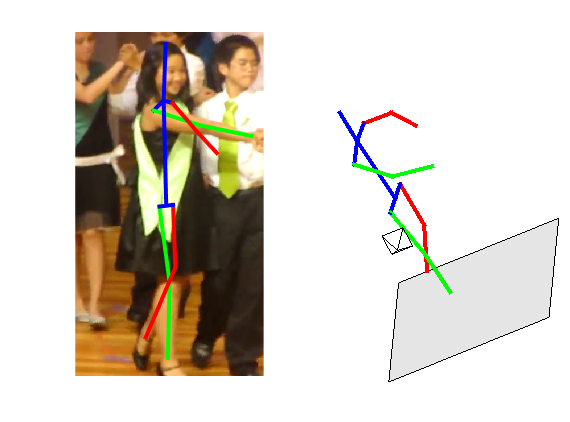}
	\includegraphics[width=0.24\linewidth,trim={2.5cm .7cm 0 .7cm},clip]{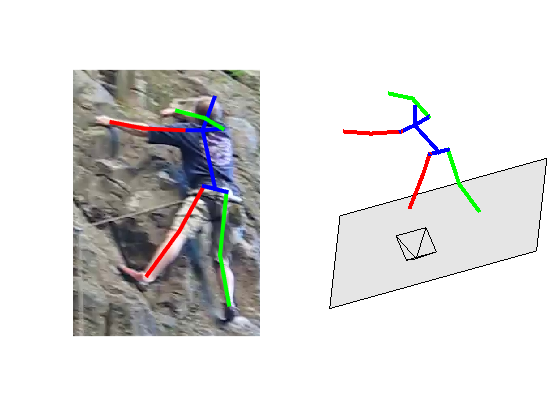}	
	\includegraphics[width=0.24\linewidth,trim={2.5cm .7cm 0 .7cm},clip]{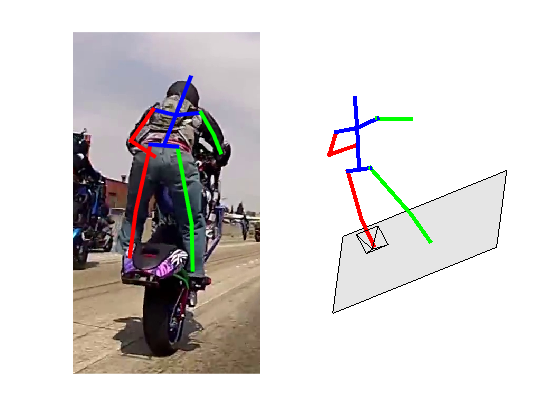}
	\includegraphics[width=0.24\linewidth,trim={2.5cm .7cm 0 .7cm},clip]{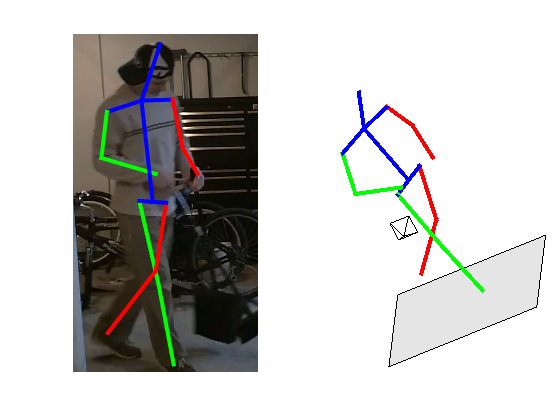}\\

	\includegraphics[width=0.24\linewidth,trim={2.5cm .7cm 0 .7cm},clip]{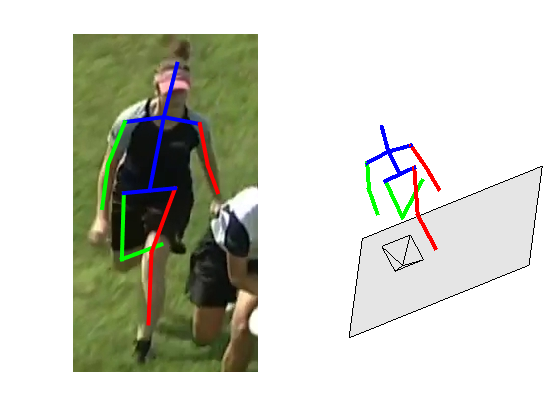}
	\includegraphics[width=0.24\linewidth,trim={2.5cm .7cm 0 .7cm},clip]{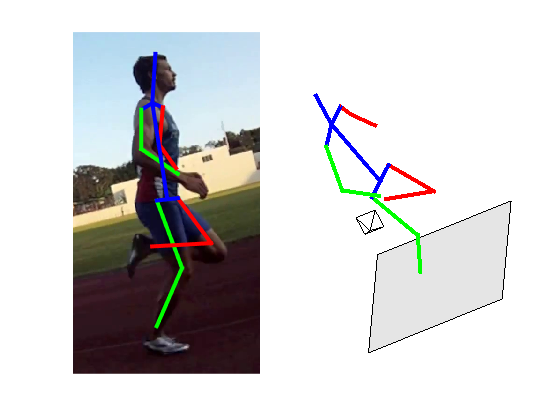}	
	\includegraphics[width=0.24\linewidth,trim={2.5cm .7cm 0 .7cm},clip]{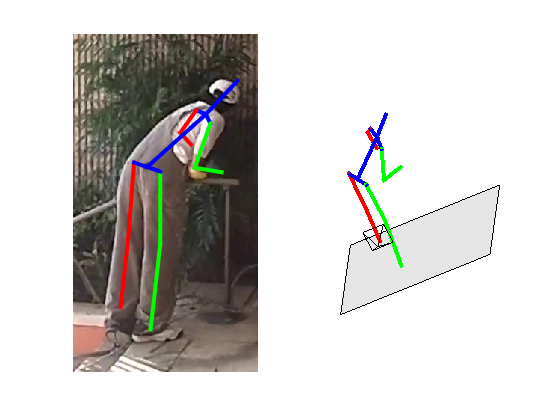}
	\includegraphics[width=0.24\linewidth,trim={2.5cm .7cm 0 .7cm},clip]{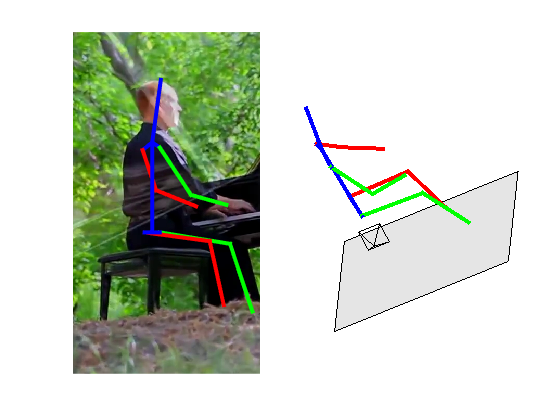}\\
	
	\includegraphics[width=0.24\linewidth,trim={2.5cm .7cm 0 .7cm},clip]{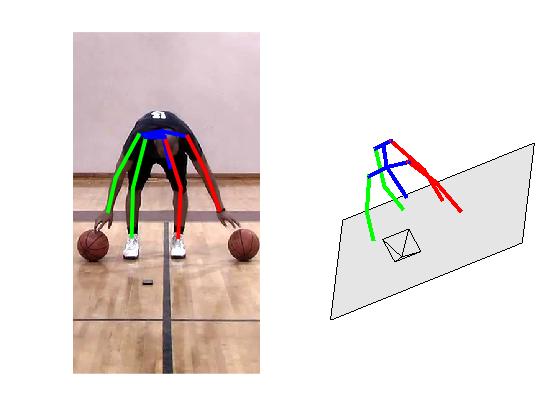}
	\includegraphics[width=0.24\linewidth,trim={2.5cm .7cm 0 .7cm},clip]{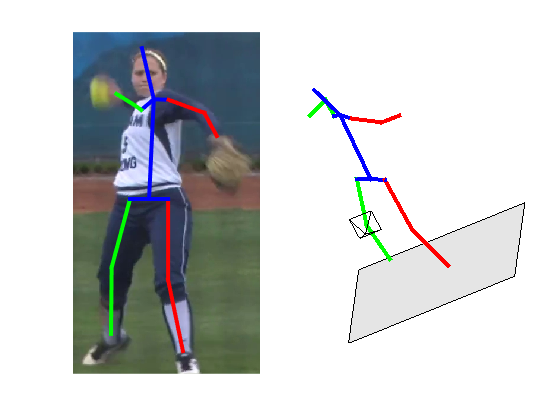}	
	\includegraphics[width=0.24\linewidth,trim={2.5cm .7cm 0 .7cm},clip]{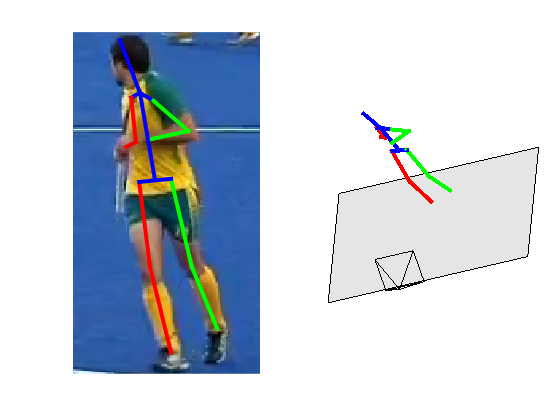}
	\includegraphics[width=0.24\linewidth,trim={2.5cm .7cm 0 .7cm},clip]{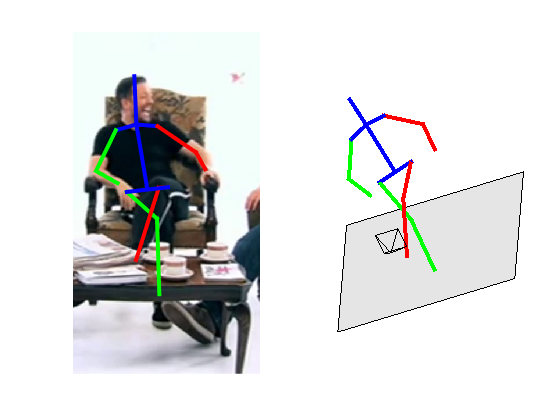}\\
	
	\includegraphics[width=0.24\linewidth,trim={2.5cm .7cm 0 .7cm},clip]{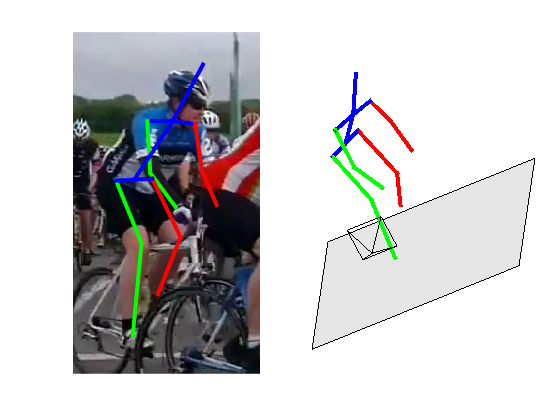}
	\includegraphics[width=0.24\linewidth,trim={2.5cm .7cm 0 .7cm},clip]{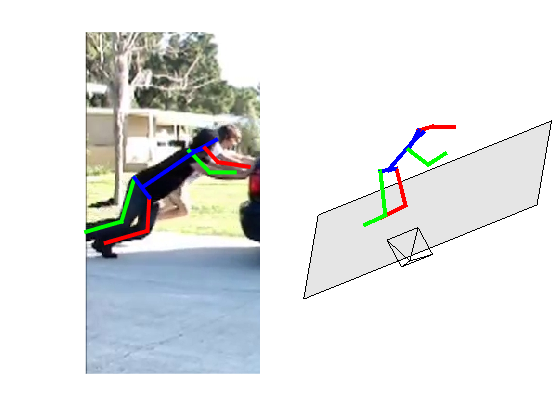}	
	\includegraphics[width=0.24\linewidth,trim={2.5cm .7cm 0 .7cm},clip]{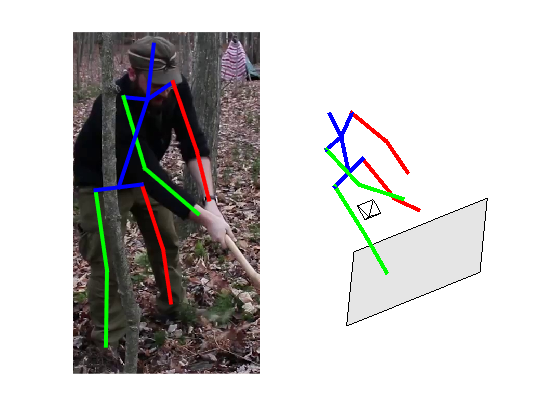}
	\includegraphics[width=0.24\linewidth,trim={2.5cm .7cm 0 .7cm},clip]{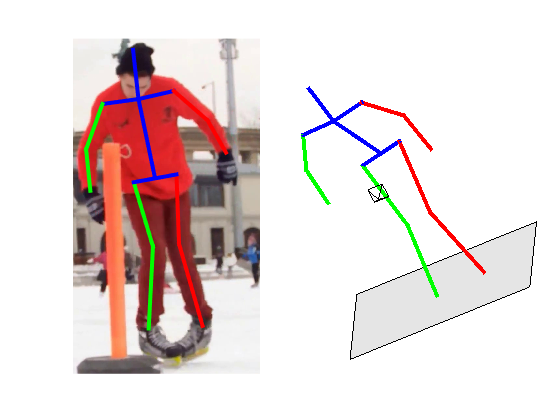}
	\\
	\caption{Examples of 3D pose reconstruction on MPII dataset. For each image we overlay 2D pose points, followed by the predicted 3D skeleton in color. The dataset does not contain ground truth 3D skeletons}
	\label{fig:mpii}
\end{figure}

\begin{figure}[htb!]
	\centering
	\includegraphics[width=0.24\linewidth,trim={2.5cm .7cm 0 .7cm},clip]{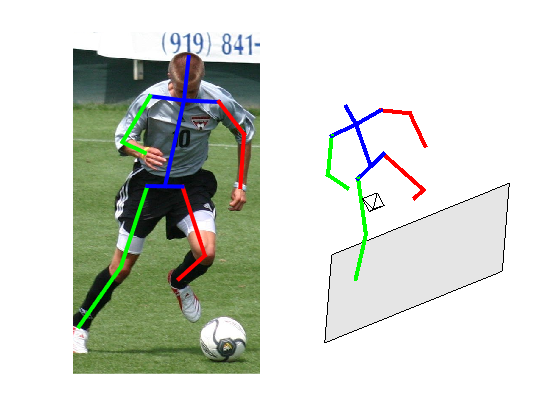}
	\includegraphics[width=0.24\linewidth,trim={2.5cm .7cm 0 .7cm},clip]{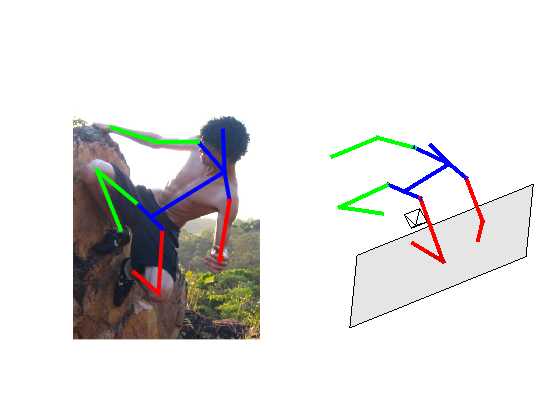}
	\includegraphics[width=0.24\linewidth,trim={2.5cm .7cm 0 .7cm},clip]{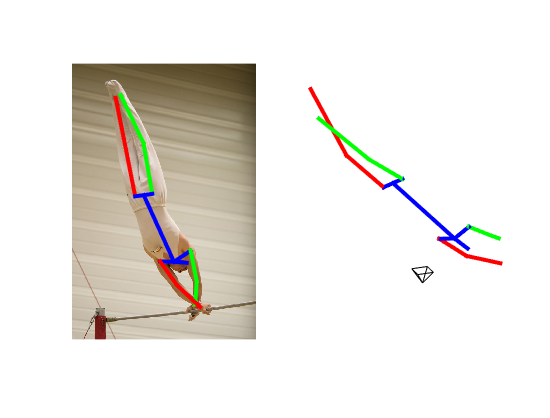}
	\includegraphics[width=0.24\linewidth,trim={2.5cm .7cm 0 .7cm},clip]{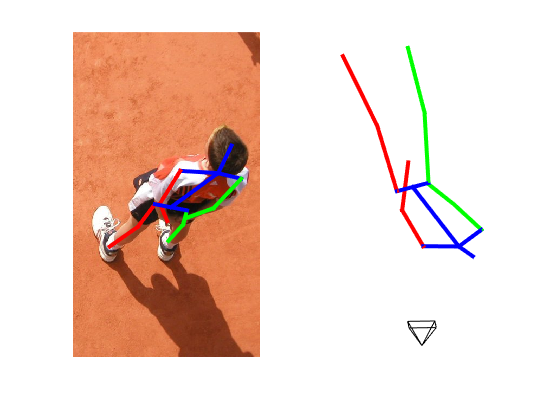}\\
	
	\includegraphics[width=0.24\linewidth,trim={2.5cm .7cm 0 .7cm},clip]{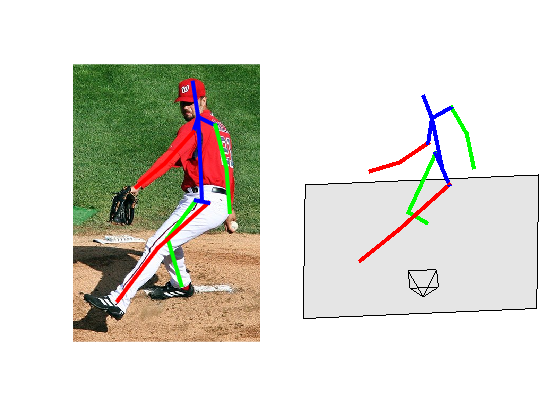}
	\includegraphics[width=0.24\linewidth,trim={2.5cm .7cm 0 .7cm},clip]{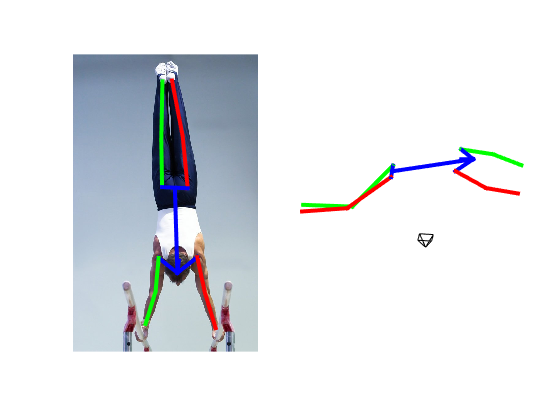}
	\includegraphics[width=0.24\linewidth,trim={2.5cm .7cm 0 .7cm},clip]{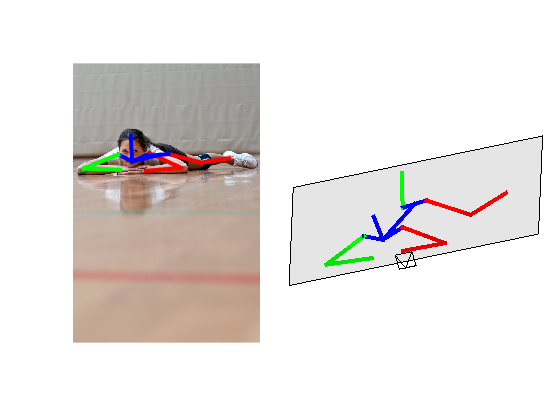}
	\includegraphics[width=0.24\linewidth,trim={2.5cm .7cm 0 .7cm},clip]{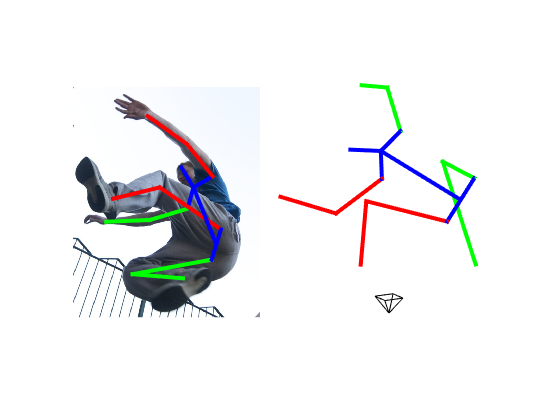}
			
	\caption{Examples of 3D pose reconstruction on Leeds dataset. For each image we overlay 2D pose points, followed by the predicted 3D skeleton in color. The dataset does not contain ground truth 3D skeletons}
	\label{fig:leeds}
\end{figure}

To test the generalization performance of our method on images in the wild, we applied our method to MPII~\cite{andriluka14cvpr} and the Leeds Sports Pose (LSP)~\cite{johnson2010clustered} datasets. MPII consists of images from short Youtube videos and has been used as a standard dataset for 2D human pose estimation. Similarly LSP dataset contains images from Flickr containing people performing sport activities. Figures~\ref{fig:mpii} and~\ref{fig:leeds} show some representative examples from these datasets containing humans in natural and difficult poses. Despite the change in domain, our weakly supervised method successfully recovers 3D poses. Note, our model is not trained on 2D poses from MPII or LSP datasets. This demonstrates the ability of our method to generalize over characteristics such as object distance, camera parameters, and unseen poses.

\subsection{Discussion}
\label{subsect:discussions}

As a general observation, we noticed that the results for the \textit{SitDown} class were the worst across the board for all methods on Human3.6M dataset. In addition to the obvious explanation of fewer available examples in this class, sit down poses lead to significant occlusion of the MoCap markers on legs or ankles (see for example Figure~\ref{fig:h3failure}).
This phenomenon leads to some of the high errors for this class.

Overall our qualitative and quantitative results have substantiated the effectiveness of using adversarial training paradigm for learning 3D priors in a weakly supervised way. Not only do we outperform the majority of similar 2D-3D lifting methods in benchmarks, we have also shown robust performance on ``images in the wild'' datasets. Even though we do not leverage any temporal information when available, we found the results from our approach to be stable when applied on per frame basis to video sequences. We share some of our (unfiltered) qualitative results on video sequences in accompanying supplementary material.

In summary, we believe that we have pushed the state of art in 3D pose estimation using weakly supervised learning. This paves way for new research directions that can extend this work by combining supervised, weakly-supervised, and unsupervised frameworks (\ie, semi-supervised) for lifting 2D poses to 3D skeletons.

\section{Conclusions}
\label{sect:conclusions}

While most of the recent progress in deep learning is fueled by labeled data, it is difficult to obtain high quality annotations for most computer vision problems. For 3D human pose estimation, acquiring 3D MoCap data remains an expensive and challenging endeavor. Our paper demonstrates that an effective prior on 3D structure {and pose} can be learned from random 2D projections {using an adversarial training framework}. 

We believe that our paper presents a unique insight into learning 3D priors via projective geometry and opens up numerous interesting applications, beyond human pose estimation. Applications such as sparse 3D reconstruction for indoor scenes as well as outdoor navigation typically requires a 3D sensor or multi-view images with a structure from motion pipeline. We envision that our approach can be applied for learning sparse 3D reconstructions from edges or keypoints extracted from a single image, using a collection of 2D images. Since the inference only requires running a small neural network, our approach can also be used for interactive graphics for rendering 3D models from line drawings, where the 3D prior is learned from a collection of such line drawings. We anticipate that our paper will spark further interest in applications combining learning techniques with projective geometry.

Finally, inspired by the results presented in this paper, we expect future work to explore the use of solely images ``in the wild'' for training the system. This could be achieved through an end to end, image to 3D adversarial pipeline or the use of a large annotated 2D pose dataset. Additional research could include analysis of difficult poses not generally seen (hand-stands, gymnast or yoga poses) as well as the use of additional datasets such as the MPI-INF-3DHP \cite{mehta2017monocular}. Improving the robustness of the system by incorporating compensation for noisy or imperfect 2D pose inputs or predicting a distribution of joint locations is another future research opportunity. There is also promise for our method to improve the state of the art by combining with fully supervised approaches.

\clearpage

\bibliographystyle{splncs04}
\bibliography{references,poseRef,ching}

\end{document}